\documentclass[sigconf]{acmart}

\AtBeginDocument{%
  }

\setcopyright{acmlicensed}
\copyrightyear{2018}
\acmYear{2018}
\acmDOI{XXXXXXX.XXXXXXX}

\acmConference[Conference acronym 'XX]{Make sure to enter the correct
  conference title from your rights confirmation emai}{June 03--05,
  2018}{Woodstock, NY}
\acmISBN{978-1-4503-XXXX-X/18/06}

\usepackage{subfigure}
\newtheorem{definition}{Definition}
\newtheorem{problem}{Problem}

\begin{document}

\title{Labor Migration Modeling through Large-scale Job Query Data}

\author{Zhuoning Guo}
\email{zguo772@connect.hkust-gz.edu.cn}
\affiliation{%
  \institution{The Hong Kong University of Science and Technology (Guangzhou)}
  \country{}
}

\author{Le Zhang}
\email{zhangle0202@gmail.com}
\affiliation{%
  \institution{Baidu Research, Baidu Inc.}
  \country{}
}

\author{Hengshu Zhu}
\email{zhuhengshu@gmail.com}
\affiliation{%
  \institution{Career Science Lab, BOSS Zhipin}
  \country{}
}

\author{Weijia Zhang}
\email{wzhang411@connect.hkust-gz.edu.cn}
\affiliation{%
  \institution{The Hong Kong University of Science and Technology (Guangzhou)}
  \country{}
}

\author{Hui Xiong}
\email{xionghui@ust.hk}
\affiliation{%
  \institution{The Hong Kong University of Science and Technology (Guangzhou)} 
  \institution{The Hong Kong University of Science and Technology}
  \country{}
}

\author{Hao Liu}
\email{liuh@ust.hk}
\affiliation{%
  \institution{The Hong Kong University of Science and Technology (Guangzhou)}
  \institution{The Hong Kong University of Science and Technology}
  \country{}
}
\renewcommand{\shortauthors}{Guo et al.}

\begin{abstract}
Accurate and timely modeling of labor migration is crucial for various urban governance and commercial tasks, such as local policy-making and business site selection. However, existing studies on labor migration largely rely on limited survey data with statistical methods, which fail to deliver timely and fine-grained insights for time-varying regional trends. To this end, we propose a deep learning-based spatial-temporal labor migration analysis framework, DHG-SIL, by leveraging large-scale job query data. Specifically, we first acquire labor migration intention as a proxy of labor migration via job queries from one of the world's largest search engines. Then, a Disprepant Homophily co-preserved Graph Convolutional Network (DH-GCN) and an interpretable temporal module are respectively proposed to capture cross-city and sequential labor migration dependencies. Besides, we introduce four interpretable variables to quantify city migration properties, which are co-optimized with city representations via tailor-designed contrastive losses. Extensive experiments on three real-world datasets demonstrate the superiority of our DHG-SIL. Notably, DHG-SIL has been deployed as a core component of a cooperative partner's intelligent human resource system, and the system supported a series of city talent attraction reports.
\end{abstract}



\maketitle

\section{Introduction}

Recent years have witnessed an increasing research interest in labor migration~\cite{stark1985new,fang2017urban,guo2022talent}. With the help of the labor migration analysis, governments can make prospective policies to retain and attract talent, and companies can select appropriate workplaces for business expansion~\cite{zhang2024urban,chao2024cross}. Nevertheless, existing studies for labor migration analysis generally rely on manually-collected survey data~\cite{martin2008managing} and the statistical methods~\cite{mamertino2016online}, which suffer from insufficient data and limited accuracy, and thus fail to provide timely and fine-grained insights for time-varying regional labor migration trends.
As a result, practical data-driven approaches based on large-scale labor migration data are desired for more effective labor migration dependencies discovery and indicators quantification.

\begin{figure}[t]
    \centering
    \includegraphics[width=\linewidth]{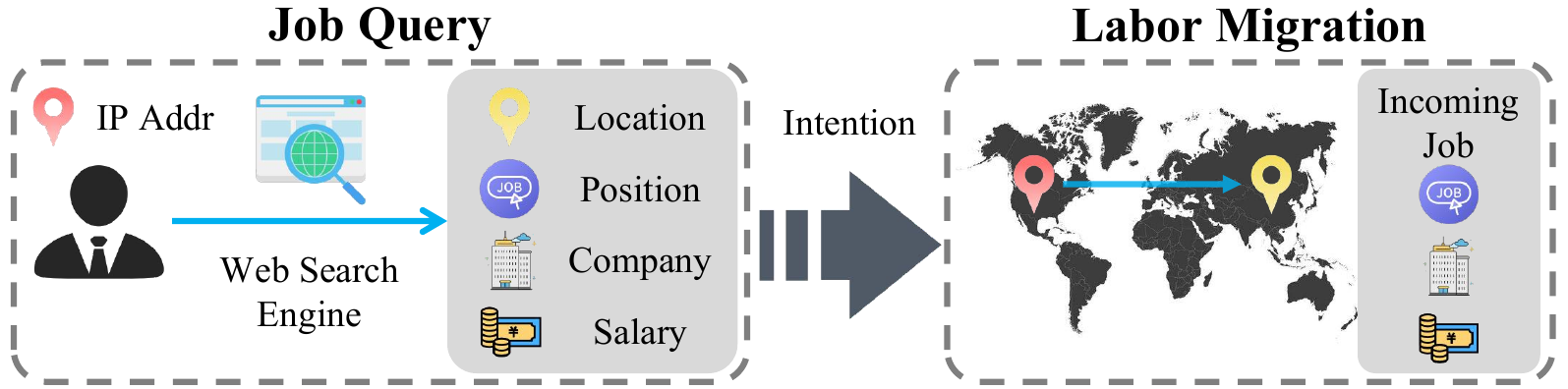}
    \caption{Job query reflects the intention of labor migration.}
    \label{fig:concept}
\end{figure}

Indeed, real-world data on labor migration is hard to access due to privacy and economic cost concerns~\cite{de2013unique,tjaden2021measuring}. Fortunately, the popularity of search engines enables the accumulation of massive job-seeking queries, reflecting users' migration intention along with the origin and destination geographies. Previous studies~\cite{lin2019forecasting,perrotta2022openness,chancellor2018measuring} have uncovered that migration intention usually presents a foresight for actual migration and can be regarded as an estimation of the latter migration action, as depicted in Figure~\ref{fig:concept}. Therefore, the intention data extracted from job-seeking queries provide unprecedented opportunities for fine-grained labor migration analysis.
In this work, we aim to provide an alternative solution for labor migration modeling based on large-scale job query data.
Specifically, we first extract cross-city labor migration intentions based on $129$ million filtered job queries from one of the most popular Chinese search engines over $350$ cities in China. 
By leveraging the large-scale labor migration data, we investigate cross-city labor migration through predictive migration modeling.

\begin{figure*}[t]
    \centering
    \subfigure[Intention-Distance Correlation.] {\includegraphics[width=0.22\linewidth]{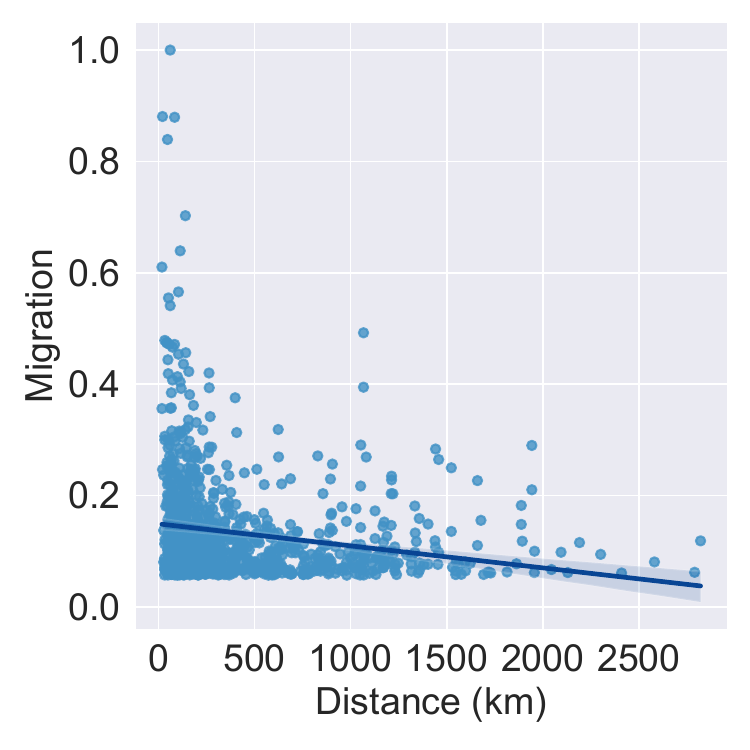}\label{fig:num_distance}}
    \subfigure[Inflow-Outflow Correlation.] {\includegraphics[width=0.22\linewidth]{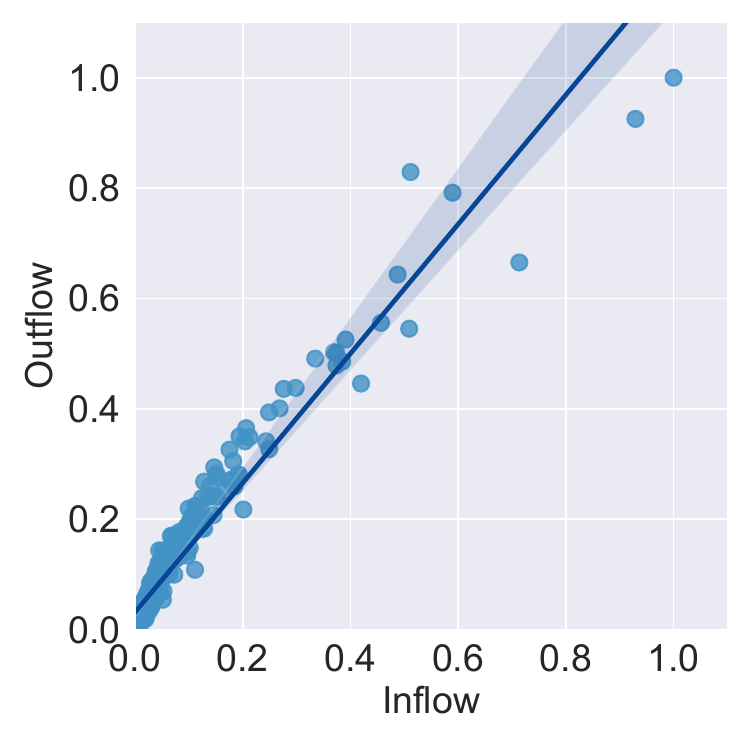}\label{fig:in_out_flow}}
    \subfigure[Labor migration intention inflow distribution of PRD.] {\includegraphics[width=0.26\linewidth]{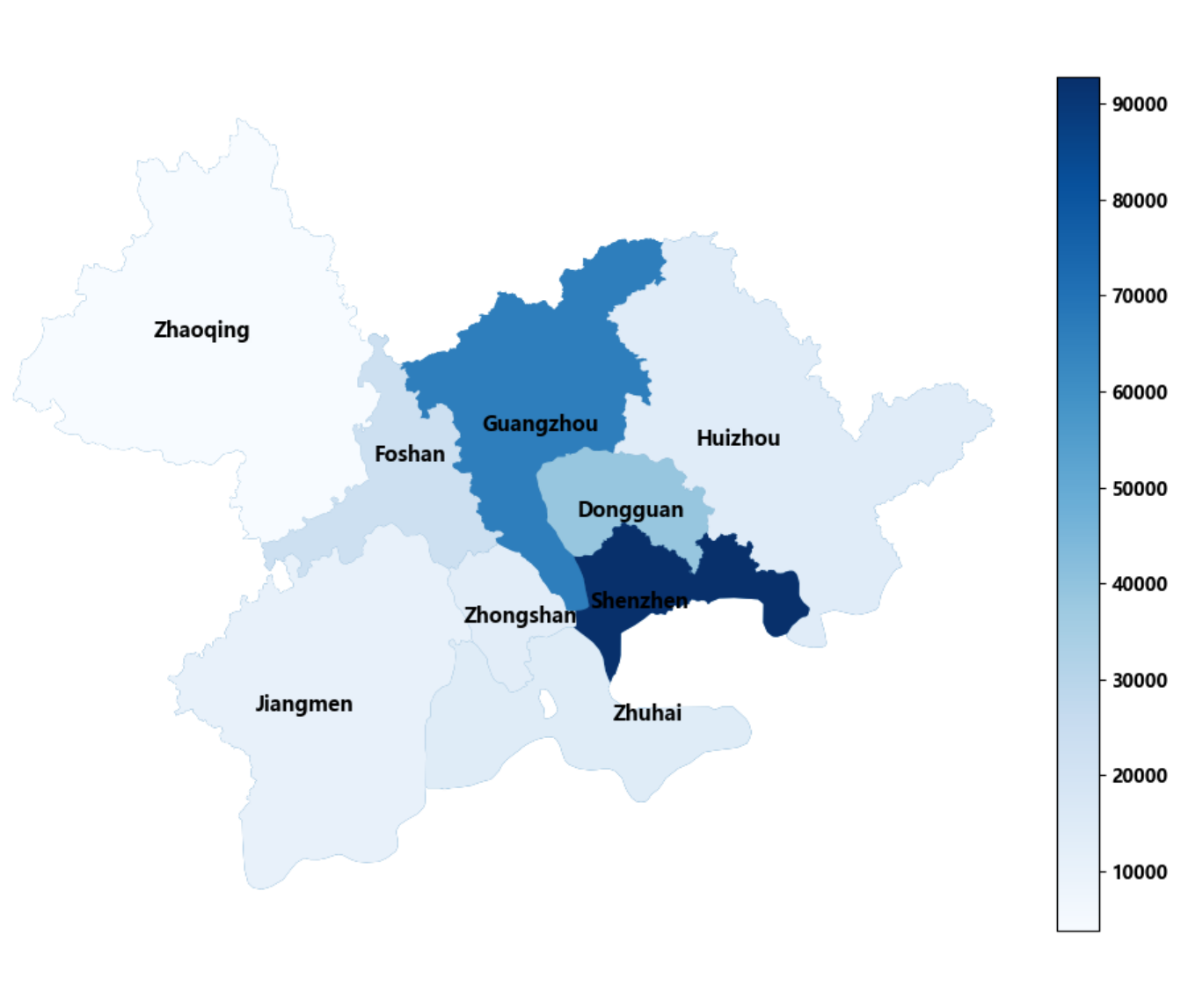}\label{fig:inflow}}
    \subfigure[Labor migration intention outflow distribution of PRD.] {\includegraphics[width=0.26\linewidth]{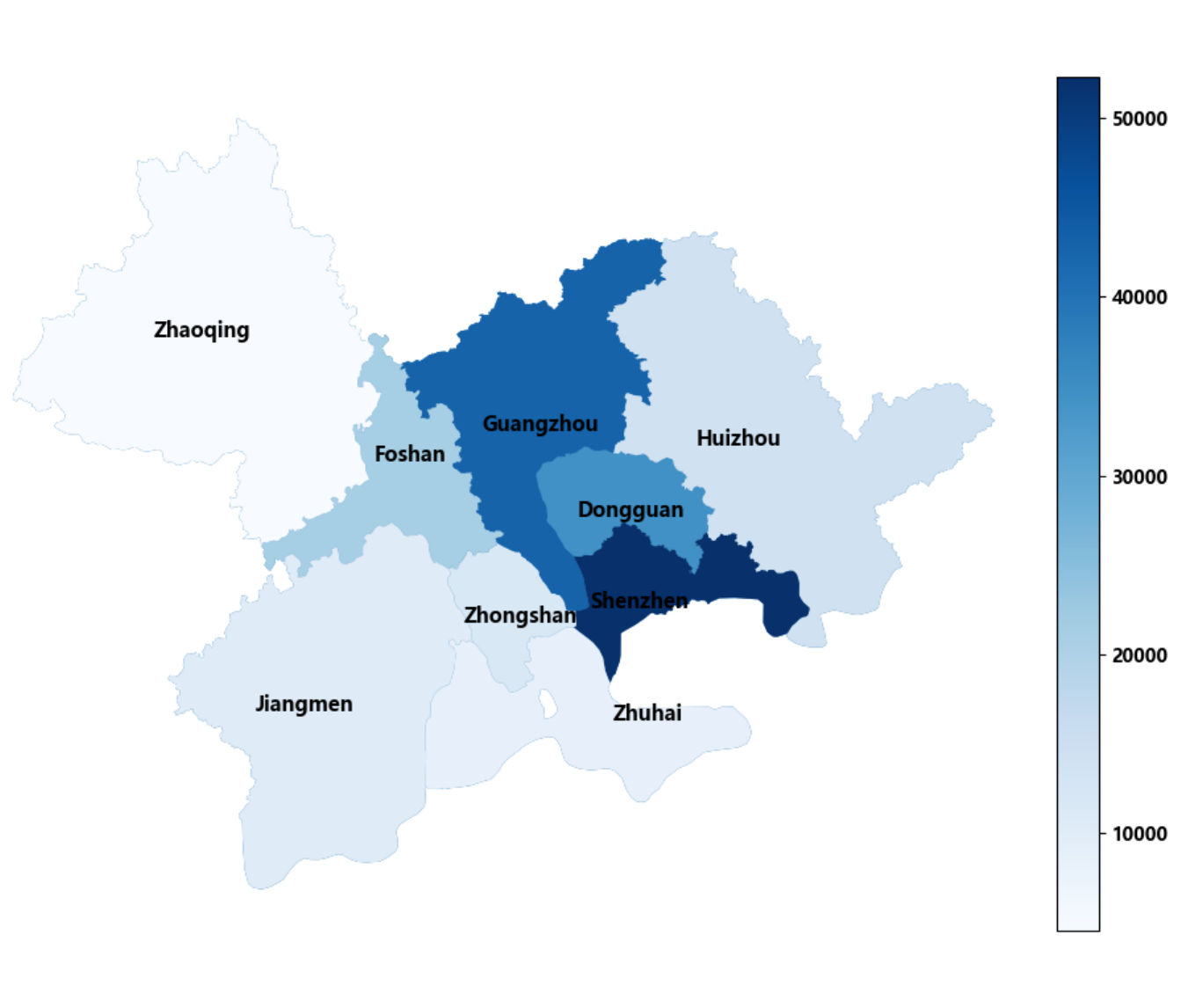}\label{fig:outflow}}
    \caption{Data statistical results on labor migration intention data.}
    \label{fig:data_exploration}
\end{figure*}

However, predicting future labor migration is a non-trivial problem because of two challenges:
~\textit{1)~Multiple discrepant homophily between cities.}
Intuitively, location and mobility patterns are two key factors significantly influencing labor migration. On the one hand, geographically closed cities may show similar talent attraction and loss patterns. On the other hand, cities with consistent labor mobility patterns in the past tend to keep similar patterns in the future. Both factors can help us to model cities' relationships, but they may be discrepant with each other. For example, Shanghai and Beijing have similar labor mobility patterns, e.g., both of them attract massive laborers from all Chinese cities, but they are distant from each other. 
Therefore, it is challenging to jointly preserve such effective but diverged relationships for labor migration modeling. 
~\textit{2)~Interpretable predictive modeling.}
The interpretability of labor migration analysis is critical to convince talents and policy-makers. Previous statistical approaches~\cite{zhao2003role,cornelius2019labor} are naturally explainable but not accurate enough, while modern deep learning-based models are more powerful but inherently black-boxed~\cite{zhang2021survey,ismail2020benchmarking}. 
How to construct a predictive model simultaneously with superior prediction capability and interpretability for quantitative and qualitative labor migration analysis is another challenge.


To tackle the aforementioned challenges, in this paper, we propose an \textit{\textbf{D}iscrepant \textbf{H}omophily co-preserved \textbf{G}raph enhanced \textbf{S}equential \textbf{I}nterpretable \textbf{L}earning}~(\textbf{DHG-SIL}) framework. Specifically, we propose a \textit{Discrepant Homophily co-preserved Graph Convolutional Network}~(DH-GCN) to learn effective city representations from two homophily-discrepant perspectives, i.e., geography and mobility. Afterward, we construct four domain-specific interpretable variables to quantify city migration properties in labor markets, which are co-optimized with city representation to absorb knowledge from the overall regression task.
Moreover, a temporal module is devised to incorporate sequential patterns of migration flows. Finally, we introduce two tailored contrastive losses to further enhance the model's interpretability by exploiting hidden supervision signals.

Our contributions include:
(1)~We investigate web-search job queries as an effective proxy for labor migration analysis.
(2)~We propose DHG-SIL, an effective framework for labor migration modeling by co-preserving cross-city and sequential labor migration dependencies.
(3)~We introduce four tailor-designed variables corresponding to cities' different labor migration properties to enhance the interpretability of the black-boxed deep learning model.
(4)~Extensive quantitative and qualitative experiments on three real-world datasets demonstrate the outperformance of DHG-SIL against baselines for a better understanding of city labor migration. The proposed framework has been deployed in our partners' intelligent report systems to provide timely insights into regional labor migration trends.

\section{Preliminaries}

In this section, we will give the preliminaries of our work, including (1)~descriptions of privacy-preserving data collection and the following pre-processing for modeling, (2)~exploring data properties for real-world findings on processed data, and (3)~important definitions of our methodology for labor migration modeling.

\subsection{Data Description}\label{sec:data_description}
We gather data on job-seeking inquiries from one of the most prominent search engines in China. This particular dataset serves as a valuable resource for gaining insight into the true statistics of labor migration, offering a unique perspective to understand factual labor migration. We outline the specifics of our data collection and pre-processing techniques in the following subsections.

\subsubsection{Privacy-preserving data collection}
In order to extract labor migration intention, we commence by gathering an extensive corpus of user queries from the search engine. For every individual query, we employ a methodology that enables us to derive a coarse-grained location, specifically at the prefecture-level city, based on the IP address and the unprocessed content of the query. It is pertinent to note that we do not have access to confidential user profiles, thus, our analysis is limited to the information that is publicly available. As a result, we are unable to ascertain precise and accurate user information.

\subsubsection{Data pre-processing}
We conditionally select queries containing job-seeking keywords (pre-processed by domain experts) and remove duplicated queries (representing identical migrating intention, i.e., with the same IP address and job-seeking keywords in a single time step). After this, $129,436,288$ job queries among $358$ cities are preserved. Secondly, based on preserved queries, we extract the search source city (where the query is issued), target city (words about cities in the query), and search timestamp. Thirdly, we aggregate and normalize preserved data samples by time step and source-target city pair as a timely pairwise cross-city dataset recording normalized numbers of laborers from a source city to a target city. 
In this way, we obtain a cross-city labor migration dataset for analysis.

\subsection{Data Exploration}\label{sec:data_exploration}
This part elucidates the analysis of the labor migration dataset to uncover data characteristics that motivate our model design.

\subsubsection{Relationship between migration and distance}
The correlation between inter-city labor migration intentions and geographical distances is reported in Figure~\ref{fig:num_distance}, illustrating the top $1\%$ of city pairs with the highest migration intentions. The results reveal a trend where greater distances correspond with fewer migration intentions, likely due to increased moving costs and homesick emotions.

\subsubsection{Correlation between inflow and outflow}
This investigation seeks to confirm a positive correlation between cities' intention migration inflows and outflows. Figure~\ref{fig:in_out_flow} suggests that cities with high labor attraction also have high turnover, as supported by previous work~\cite{card2001immigrant}, indicating that dynamic labor markets foster both high entry and exit rates.

\subsubsection{Discrepancy of city homophily}
Taking the Pearl River Delta Urban Agglomeration for example, we visualize the migration inflow and outflow in Figure~\ref{fig:inflow} and Figure~\ref{fig:outflow}, respectively. We observe that cities with similar intention values are not always geographically close, demonstrating that the similarities between cities in a geography view and those in a mobility view are not usually consistent. This phenomenon motivates us to capture diverged city correlations for better migration modeling.

\begin{figure*}[t]
	\centering
	\includegraphics[width=\linewidth]{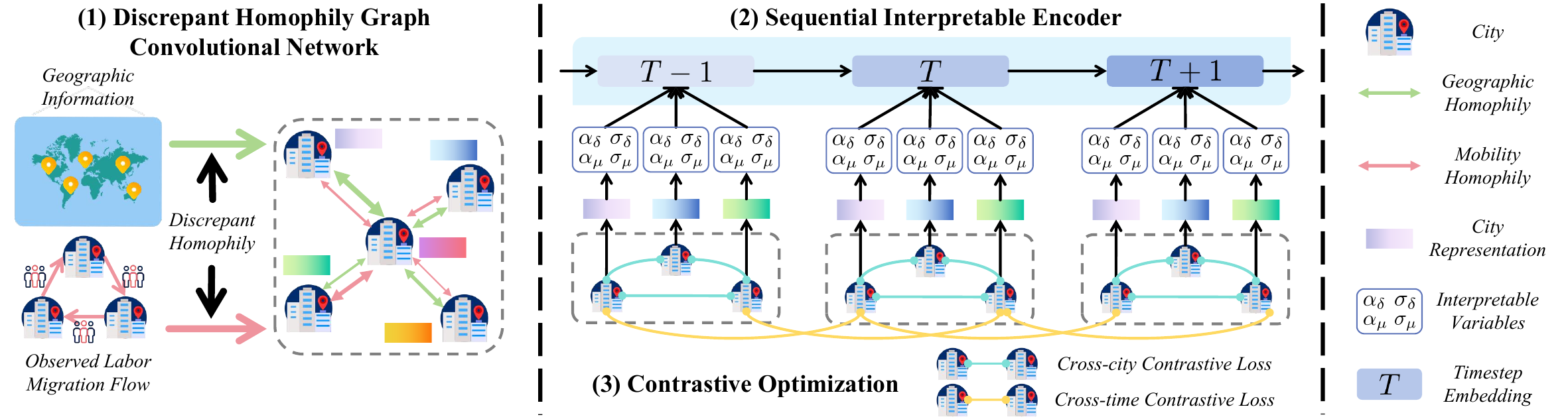}
	\caption{An overview of Discrepant Homophily co-preserved Graph enhanced Sequential Interpretable Learning framework.}
	\label{fig:sil}
\end{figure*}

\subsection{Problem Formulation}\label{sec:def}
In this subsection, we introduce the core definitions throughout this paper.
Here we first denote migration values, attraction, and repulsion.

We denote \textit{Migration Value $i_{c_1 \rightarrow c_2}$} as the number of laborers migrating from city $c_1$ to city $c_2$. Note we use the labor migration intention to approximate the factual labor migration.
\begin{definition}
    \textit{Attraction $\delta$} describes the ability of a city to attract laborers (i.e., come to the city) from other cities.
\end{definition}
\begin{definition}
    \textit{Repulsion $\mu$} describes the extent of a city to repulse laborers (i.e., leave the city) to other cities.
\end{definition}
Given two cities $c_1$ and $c_2$, there are two intuitive regularities:
\begin{itemize}
    \item $i_{c_1 \rightarrow c_2}$ is positively related to the repulsion of $c_1$ and the attraction of $c_2$;
    \item $i_{c_1 \rightarrow c_2}$ is negatively related to their distance $d_{c_1,c_2}$.
\end{itemize}

To quantify attraction and repulsion, we further define two variables: intensity describes the maximum labor ability, and attenuation incorporates geographic distance information. Here are their definitions.
\begin{definition}
    \textit{Intensity $\alpha$} is a relative value that measures how much a city attracts (repulses) laborers from (to) other cities. Without considering other factors, the number of laborers attracting (repulsing) from (to) a common target city is positively correlated to the attraction (repulsion) intensity.
\end{definition}
\begin{definition}
    \textit{Attenuation $\sigma$} is a relative value to describe the attenuation rate of migration as the distance increases. The migration value of a city is naturally lower for remote cities than close cities. In other words, keeping intensity constant, a city with higher attenuation attracts (repulses) more laborers from (to) other cities.
\end{definition}

Additionally, we define \textit{Migration Flow} as the matrix of city pairwise migration values among city set $C$, which can be formulated as
\begin{equation}
    \mathbf{I}_C = \{i_{c_1 \rightarrow c_2}|c_1, c_2 \in C\},
\end{equation}
where $c_1$ is the source city and $c_2$ is the target city.
Finally, we formulate the labor migration prediction problem as follows:
\begin{problem}\label{problem}
    Given a city set $C$ with their corresponding coordinates and historical observed cross-city labor migration flow $\mathbf{X} = \{\mathbf{I}_C^t\}_{t=1}^T$, we aim to predict the migration flow at the next time step as
    \begin{equation}
        \hat{\mathbf{I}}_C^{T+1} \gets \mathcal{F}(\mathbf{X}; C),
    \end{equation}
    where $\hat{\mathbf{I}}_C^{T+1}$ is the estimated migration flow at the time step $T+1$ among all cities $C$, and $\mathcal{F}$ is the predictive function required to learn.
\end{problem}

\section{Framework}

This section presents the DHG-SIL framework for labor migration analysis, as illustrated in Figure~\ref{fig:sil}.
First, we introduce a Discrepant Homophily co-preserved Graph Convolutional Network (DH-GCN) to extract sophisticated cross-city and cross-feature dependencies for learning effective city representation from two homophily-discrepant perspectives. Then, we construct four defined interpretable variables to quantify city migration properties under co-optimization with the predictive objective. Afterward, a temporal module is devised to capture sequential patterns of time-varying labor migration flows. Last, two tailor-designed contrastive losses are introduced to enhance the model interpretability and prediction accuracy by leveraging the hidden supervision signals.

\label{sec:dhgcn}
\subsection{Discrepant Homophily Co-preserved Graph Convolutional Network}

Two categories of city-related features in labor migration should be incorporated, including geography and mobility. We propose to quantify the following two types of pairwise city relationships.
\textbf{1)~Geography-homophilic.} According to Tobler's First Law of Geography, close cities in a shared urban agglomeration show more similar labor migration patterns, which we term geography-homophilic.
\textbf{2)~Mobility-homophilic.} Cities with similar historical labor migration flows are mobility-homophilic.
In this part, we leverage Graph Neural Network~(GNN) to learn city representations by incorporating the aforementioned homophilic features.
However, as mentioned in Section~\ref{sec:data_exploration}, the discrepancy of two types of features may violate the homophily assumption on graphs for representation learning~\cite{zheng2022graph}, i.e., nodes with similar features or consistent class labels are linked together.
To consolidate the diverged information, we introduce the \textit{Discrepant Homophily co-preserved Graph Convolutional Network}~(DH-GCN) for city representation learning.

A geography graph is first formulated below to preserve geography-homophily, where nodes represent cities and edges are determined by the geographical distance.
\begin{definition}
    \label{def:geograph}
    \textbf{Geography Graph $\mathcal{G} = (\mathcal{V}, \mathcal{E})$.} $\mathcal{V}$ corresponds to the set of city nodes $C$, and $\mathcal{E}$ is the set of edges between close city pairs.
    For any node pair $u$ and $v$, we add an edge $e_{uv} \in \mathcal{E}$ if their distance $d_{uv} \leq \epsilon$, where $\epsilon$ is a pre-defined spherical distance threshold. 
\end{definition}

Moreover, we design a metric-based graph convolution kernel for modeling mobility homophily. 
Specifically, we define the following metric to quantify the difference in migration flow for different city pairs:
\begin{equation}
    \Delta_{uv} = \sqrt{\sum\limits^{d_{uk}=d_{vk}}_{k\in K_{uv}}\frac{\Vert{I_{k \rightarrow u} - I_{k \rightarrow v}}\Vert_2 + \Vert{I_{u \rightarrow k} - I_{v \rightarrow k}}\Vert_2}{2|K_{uv}|}},
\end{equation}
where $u$, $v$ and $k$ are three cities, $K_{uv}$ is the set of cities satisfying $d_{uk}=d_{vk}$ and $|K_{uv}|$ is the set size, $d_{uk}$ is the spherical distance between $u$ and $k$, $I_{k \rightarrow u}$ is a list of migration values from $k$ to $u$, and $\Vert{I_{k \rightarrow u} - I_{k \rightarrow v}}\Vert_2$ is the 2-Norm of difference of two lists. We compare the list of migration values in previous $T$ time steps (i.e., $I_{k \rightarrow u}={[i^t_{k \rightarrow u}]}_{t=1}^T$).

\begin{table*}[t]
    \centering
    \caption{Datasets statistics. We present the data sizes and corresponding urban agglomeration information in 2022.}
    \begin{tabular}{c|cc|ccccc}
        \toprule
        Dataset & $\#$ of cities & $\#$ of job queries & Urban Agglomeration & Representative City & GDP (RMB) & Population & Area ($km^2$) \\
        \midrule
        YRD & 41 & 26,841,230 & Yangtze River Delta & Shanghai & 29.03 Trillion & 174 Million & 212 Thousand\\
        PRD & 9 & 7,816,687 & Pearl River Delta & Guangzhou & 10.05 Trillion & 57 Million & 56 Thousand\\
        BTH & 14 & 8,729,264 & Beijing-Tianjin-Hebei & Beijing & 9.6 Trillion & 114 Million & 215 Thousand\\
        \bottomrule
    \end{tabular}
    \label{tab:dataset}
\end{table*}

Then, we devise a graph convolutional network on the constructed geography graph to incorporate two types of homophiles, 
\begin{equation}
    h^{(l+1)}_u = \sum\limits_{v\in N_u} b^{(l)} + \frac{e^{-d_{uv}} \cdot \Delta_{uv}}{\sqrt{|N_u||N_v|}} \cdot h^{(l)}_v \cdot \mathbf{W}^{(l)},
\end{equation}
where $u$ and $v$ are two cities, $h^{(l+1)}_u$ is the representation of $u$ of the $l$-th GNN layer, $N_u$ denotes the set of neighbors of $u$, $d_{uv}$ is the spherical distance between $u$ and $v$, and $b^{(l)}$ and $\mathbf{W}^{(l)}$ are learnable parameters of the $l$-th layer.

However, the above graph convolution operation has two limitations. First, the mobility information is not comprehensively considered for all cities because $\Delta_{uv}$ is only calculated between connected cities.
Second, the strict condition of $K_{uv}$ is not practical in the real world and the computation is not efficient.
By regarding any two cities connected in the geography graph (i.e., $d_{uv} \leq \epsilon$) are approximately equal distant from others, we relax $d_{uk}=d_{vk}$ to $d_{uk}-d_{vk}<\tau$, and let $K_{uv}=C$. 
The feasibility can be proved based on a regularity that in a triangle $\triangle_{uvk}$, when length $\overline{uv}$ approaches zero, length $\overline{uk}$ and $\overline{vk}$ are approximately equal. 
Based on such relaxation, we revise the spectral graph convolution in a more simple and efficient form
\begin{equation}
    \begin{array}{c}
        \mathbf{H}^{(l+1)}_C = b^{(l)} + \mathbf{M}_C^t \cdot \mathbf{H}^{(l)}_C \cdot \mathbf{W}^{(l)}, \\
        \mathbf{M}_C^t = \frac{1}{2}\mathbf{A}^t_{\mathcal{G}} \cdot \mathbf{D}^t_{C} \cdot [\mathbf{I}^t_{C} \otimes {(\mathbf{I}^t_{C})}^T + {(\mathbf{I}^t_{C})}^T \otimes \mathbf{I}^t_{C}],
    \end{array}
\end{equation}
where $\mathbf{H}^{(l)}_C = [h^{(l)}_{c_1}, h^{(l)}_{c_2}, \cdots]$ is the representation matrix of $C$ of the $l$-th layer, $\mathbf{A}^t_{\mathcal{G}}$ is the adjacency matrix of geography graph $\mathcal{G}$ at time step $t$, and $\mathbf{D}^t_{C}$ is the distance matrix of $C$ at time step $t$. Besides, $\otimes$ is an operation on any two $N \times N$ matrices $\mathbf{A}\{a_{ij}\}$ and $\mathbf{B}\{b_{ij}\}$ as $\mathbf{C}\{c_{ij}\} = \mathbf{A}\{a_{ij}\} \otimes \mathbf{B}\{b_{ij}\},c_{ij} = 1 - \small{\sqrt{\sum^{k=1}_{k \leq N} \frac{{\Vert a_{ik} - b_{kj} \Vert}_2}{N}}}$.

\subsection{Sequential Interpretable Encoding}

Then, we introduce the encoding of migration flow sequences based on interpreting city representations, including two steps: 1)~interpretable static modeling of migration flow and 2)~sequential encoding of static migration flows.

\textbf{Interpretable static modeling.}
To model migration values in one time step, we propose inferring the migration flow by transforming the city representation based on interpretable variables and functions.
Based on city representations derived by DH-GCN, a Multi-Layer Perceptron (MLP) is devised to transform city representations into four variables: attraction intensity, attraction attenuation, repulsion intensity, and repulsion attenuation.
Then, to model the migration value statically, we define an interpretable function, Attraction-Repulsion based Migration Flow in below.

\begin{definition}\label{def:ARMF}
    \textbf{Attraction-Repulsion based Migration Flow (ARMF).} We define the migration value between a source city $c_1$ and a target city $c_2$ as the product of the attraction of the target city $\delta_{c_2}$ and the repulsion of the source city $\mu_{c_1}$.
    We denote $d_{c_1, c_2}$ as the distance between $c_1$ and $c_2$. Attraction distribution is defined as
    \begin{equation}
        \delta = \mathcal{F}_\delta(d_{c_1, c_2}|\alpha_\delta; \sigma_\delta) = \alpha_\delta\exp{(-d_{c_1, c_2}^2\sigma_\delta^2)},
    \end{equation}
    where $\alpha_\delta$ is the attraction intensity and $\sigma_\delta$ is the attraction attenuation.
    Repulsion distribution is defined as
    \begin{equation}
        \mu = \mathcal{F}_\mu(d_{c_1, c_2}|\alpha_\mu; \sigma_\mu) = \alpha_\mu\exp{(-d_{c_1, c_2}^2\sigma_\mu^2)},
    \end{equation}
    where $\alpha_\mu$ is repulsion intensity of and $\sigma_\mu$ is repulsion attenuation.
    Finally, the migration flow is formulated as
    \begin{equation}
        \begin{array}{rl}
            i_{c_1 \rightarrow c_2} & = \delta_{c_2} \cdot \mu_{c_1} = \mathcal{F}_\delta(d_{c_1, c_2}|\alpha_\delta; \sigma_\delta) \cdot \mathcal{F}_\mu(d_{c_1, c_2}|\alpha_\mu; \sigma_\mu) \\
                                    & = \alpha_\delta \alpha_\mu \exp{(-d_{c_1, c_2}^2\sigma_\delta^2-d_{c_1, c_2}^2\sigma_\mu^2)}.
        \end{array}
    \end{equation}
\end{definition}

ARMF is based on formulations of attraction and repulsion distributions according to the regularities.
Inspired by the Gravity Model~\cite{dudas2017virus}, we define their distributions based on Gaussian Distribution, which are then multiplied as migration flows.
The interpretable static modeling allows for inferring migration values and identifying four domain-specific variables based on real-world labor migration scenarios. Unlike existing approaches, We quantify them through co-optimization with the main regression task, jointly preserving interpretability and effectiveness.

\textbf{Sequential encoding.}
Labor migration is time-varying as the city properties evolve. After modeling static migration values between cities, to predict future migration values at the next time step based on observed ones, we adopt a popular deep learning-based time series model, \textit{N-BEATS}~\cite{oreshkin2019n} to encode the temporal pattern of time-varying migration flows.
We can describe the migration flow $I_C^{t}$ as a $|C| \times |C|$ matrix. To reduce its complexity and extract effective low-dimensional features, we use MLP to get a $|C|$-dimensional vector and calculate $|C|$ eigenvalues, which are concentrated as the feature of each time step.
The time-varying interpretable variables will be optimized with all parameters in our end-to-end model during the training process, which will be detailed in the next section.

\subsection{Contrastive Optimization}

Beyond the supervision signal of the ground-truth migration flow, we further incorporate prior knowledge as supervision signals to improve the interpretable variables' effectiveness for downstream qualitative analysis.
Specifically, we construct pairwise contrastive losses to achieve cross-city and cross-time supervision. First, within the same time step, the difference in immigration or migration values of two cities is usually positively related to the ratio of their attraction or repulsion intensity. Second, for a particular city, the difference in immigration or migration values across two consecutive time steps is also usually positively related to the ratio of attraction or repulsion intensity of the city itself.

We define a contrastive loss function $\mathcal{L}$ for cross-time and cross-city computation as
\begin{equation}
    \mathcal{L}(\rho_i;\rho_\alpha;\lambda) = \max(0,-\rho_i\cdot\rho_\alpha+\lambda),
\end{equation}
where $\rho_i$ is the difference of migration values, $\rho_\alpha$ is the difference of intensity, and $\lambda$ is a constant controlling the contrastive degree in optimization. Based on the function, we can derive cross-time contrastive loss $l_t$ as
\begin{equation}
    \begin{array}{rl}
        l_{j,(t_1,t_2)} = & \mathcal{L}(i_{C_{t_1} \rightarrow c_{j,t_1}} - i_{C_{t_2} \rightarrow c_{j,t_2}};\alpha_\delta^{c_{j,t_1}} - \alpha_\delta^{c_{j,t_2}};\lambda^{\delta}) \\
         & + \mathcal{L}(i_{c_{j,t_1} \rightarrow C_{t_1}} - i_{c_{j,t_2} \rightarrow C_{t_2}};\alpha_\mu^{c_{j,t_1}} - \alpha_\mu^{c_{j,t_2}};\lambda^{\mu}),
    \end{array}
\end{equation}
where $c_{j,t_1}$ and $c_{j,t_2}$ are $j$-th city at $t_1$ and $t_2$ time steps.
Similarly, the cross-city contrastive loss $l_c$ can be derived as
\begin{equation}
    \begin{array}{rl}
        l_{(j_1,j_2),t} = & \mathcal{L}(i_{C_{t} \rightarrow c_{j_1,t}} - i_{C_{t} \rightarrow c_{j_2,t}};\alpha_\delta^{c_{j_1,t}} - \alpha_\delta^{c_{j_2,t}};\lambda^{\delta}) \\
         & + \mathcal{L}(i_{c_{j_1,t} \rightarrow C_{t}} - i_{c_{j_2,t} \rightarrow C_{t}};\alpha_\mu^{c_{j_1,t}} - \alpha_\mu^{c_{j_2,t}};\lambda^{\mu}),
    \end{array}
\end{equation}
where $c_{j_1,t}$ and $c_{j_2,t}$ are $j_1$-th and $j_2$-th cities at $t$ time step.

The overall objective function can be defined as
\begin{equation}
    \begin{array}{rl}
        l = & \sum^{t=1}_{T}({\Vert \hat{I_C^{t}} - I_C^{t} \Vert}_2) \\
        & + \sum^{j_1=1}_{|C|-1}\sum^{j_2=j_1+1}_{|C|}\sum^{t=1}_{T}l_{(j_1,j_2),t} \\
        & + \sum^{t_1=1}_{T-1}\sum^{t_2=t_1+1}_{T}\sum^{j=1}_{|C|}l_{j,(t_1,t_2)},
    \end{array}
\end{equation}
where $l$ is the overall loss value, and $\{\hat{I_C^{t}}\}_{t=1}^T$ and $\{I_C^{t}\}_{t=1}^T$ are the sets of estimated and ground-truth migration flows.

\label{sec:experiments}
\section{Experiments}

\noindent
\textbf{Experimental datasets.}
We construct three datasets from the original collected data, namely \textit{YRD}, \textit{PRD}, and \textit{BTH}, from the three largest urban agglomerations~\cite{fang2017urban} that are \textit{Yangtze River Delta Urban Agglomerations}, \textit{Pearl River Delta Urban Agglomerations} and \textit{Beijing-Tianjin-Hebei Urban Agglomerations}, respectively. The time ranges of the three datasets are all from Jan. 2020 to Dec. 2021 in a monthly granularity. More statistics and information are listed in Table~\ref{tab:dataset}.

\noindent
\textbf{Evaluation metrics.}
As a regression task, we adopt the widely used Mean Absolute Error~(MAE) and Root Mean Square Error~(RMSE) to evaluate the overall performance.

\noindent
\textbf{Baseline models.}
We compare DHG-SIL with several baseline methods, including
\textit{Mean}, 
\textit{LR}, 
\textit{GBRT}~\cite{friedman2001greedy}, 
\textit{LSTM}~\cite{hochreiter1997long}, 
\textit{TF}~(Transformer)~\cite{vaswani2017attention}, 
\textit{STGCN}~\cite{yu2017spatio}, 
\textit{DCRNN}~\cite{li2017diffusion}, 
and \textit{Ahead}~\cite{zhang2021attentive}.

\noindent
\textbf{Implementation details.}
For DHG-SIL, we choose parameters including the length of labor migration time steps $T=5$, the max geo-homophilic distance of $\epsilon=100km$ for YRD and PRD, $200km$ for BTH, the embedding dim of graph node representation as $8$, the contrastive learning margin $\lambda^{\delta} = \lambda^{\mu} = 0.1$, the N-BEATS width as $256$, the N-BEATS expansion coefficient length as $32$ and the three blocks of N-BEATS are \textit{generic}, \textit{seasonality} and \textit{trend}~\cite{oreshkin2019n}.
We leverage the Adam~\cite{kingma2014adam} to optimize the end-to-end model with learning rate as $0.001$.
The DHG-SIL is run on the machine with Intel Xeon Gold 6148 @ 2.40GHz, V100 GPU, and 64G memory.

\subsection{Overall Results}

Table~\ref{tab:overall} reports the overall experimental results of model performance with a comparison of baselines on three datasets (i.e., \textit{YRD}, \textit{PRD}, and \textit{BTH}) regarding two metrics (i.e., MAE and RMSE). We observe that DHG-SIL achieves the best performance (i.e., lowest MAE and RMSE) among all baselines, and obtains a significant improvement (at least $30.16\%$, $36.60\%$, and $22.34\%$ for MAE, and $24.41\%$, $26.42\%$ and $17.92\%$ for RMSE) on three datasets. First, the outperformance of DHG-SIL compared with LSTM demonstrates that we extract diverse city representations that provide rich information for sequential encoding. Second, DHG-SIL also defeats STGCN and DCRNN in labor migration tasks, which shows that our specific feature extraction methods effectively pave the way for precise optimization for prediction. Last, compared with Ahead, a job mobility prediction model, DHG-SIL takes more consideration of geographical properties and the complex relationship between cities in multiple perspectives, which results in lower errors in regression tasks.

\begin{table}[t]
    \centering
    \caption{The overall performance of labor migration prediction on three real-world datasets.}
    \label{tab:overall}
    \begin{tabular}{c|cc|cc|cc}
        \toprule
        & \multicolumn{2}{c|}{YRD} & \multicolumn{2}{c|}{PRD} & \multicolumn{2}{c}{BTH}\\
        \midrule
        & \footnotesize{MAE} & \footnotesize{RMSE} & \footnotesize{MAE} & \footnotesize{RMSE} & \footnotesize{MAE} & \footnotesize{RMSE}\\
        \midrule
        Mean  & 1.332 & 2.670 & 1.627 & 5.281 & 1.430 & 3.605 \\
        LR & 0.684 & 0.738 & 0.693 & 0.885 & 0.645 & 0.794 \\
        GBRT  & 0.458 & 0.578 & 0.593 & 0.829 & 0.517 & 0.682 \\
        LSTM  & 0.374 & 0.488 & 0.434 & 0.603 & 0.357 & 0.497 \\
        TF & 0.305 & 0.357 & 0.293 & 0.330 & 0.273 & 0.340 \\
        STGCN & 0.296 & 0.358 & 0.279 & 0.318 & 0.233 & 0.281 \\
        DCRNN & 0.270 & 0.340 & 0.205 & 0.259 & 0.205 & 0.268 \\
        Ahead & 0.265 & 0.336 & 0.227 & 0.279 & 0.204 & 0.268 \\
        \midrule
        DHG-SIL & \textbf{0.185} & \textbf{0.254} & \textbf{0.130} & \textbf{0.191} & \textbf{0.158} & \textbf{0.220} \\
        \bottomrule
    \end{tabular}
\end{table}

\begin{figure}[t]
    \centering
    \subfigure[MAE] {\includegraphics[width=.47\linewidth]{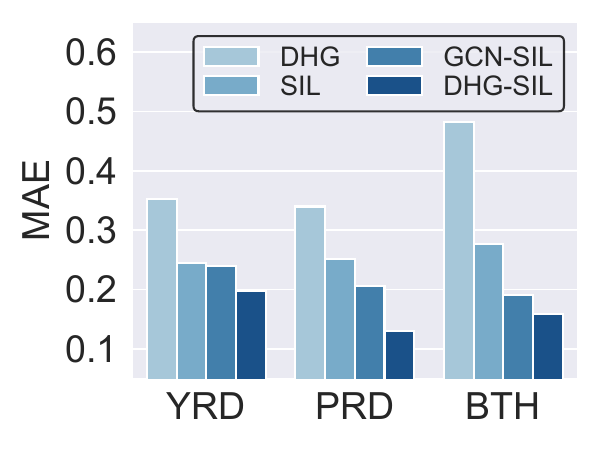}}
    \hspace{3mm}
    \subfigure[RMSE] {\includegraphics[width=.47\linewidth]{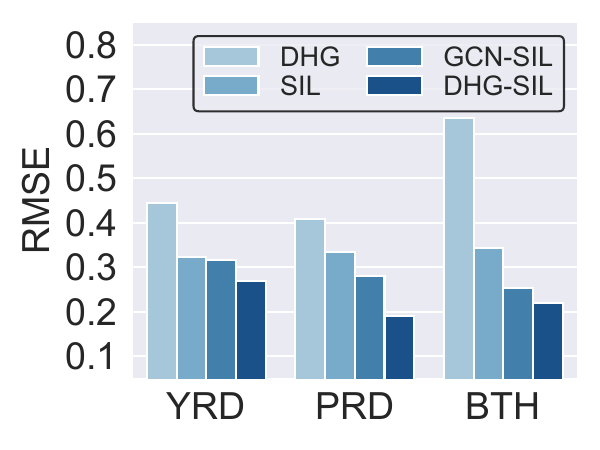}}
    \caption{Ablation study results of three datasets.}
    \label{fig:ablation}
\end{figure}

\subsection{Ablation Study}

An ablation study was conducted to assess the performance of DHG-SIL and its variants across three datasets. The variants included: 1)~DHG, which omits the sequential interpretable learning modules and relies solely on DH-GCN for static migration flow prediction, averaging encoded observations; 2)~SIL, which excludes the DH-GCN component, using only the sequential module for migration flow prediction; and 3)~GCN-SIL, which replaces DH-GCN with a standard GCN~\cite{kipf2016semi}.
Results depicted in Figure~\ref{fig:ablation} indicate that evidenced by increased MAE and RMSE, the removal or substitution of modules significantly diminishes predictive accuracy. Notably, the DHG variant underperforms due to its inability to capture sequential dynamics, while SIL is suboptimal because it neglects geographic and mobility relationships. GCN-SIL is also less effective than DHG-SIL, as DH-GCN is specifically tailored for the inherent discrepant homophily, unlike the more generic GCN designed for homophilic graphs.

\begin{figure}[t]
	\centering
	\subfigure[Intensity] {\includegraphics[width=.47\linewidth]{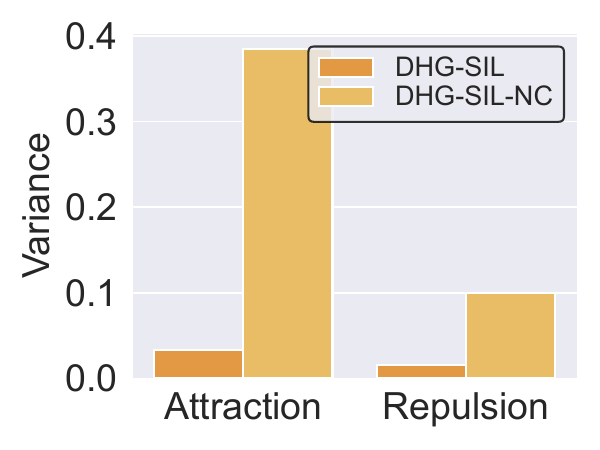}}
 \hspace{3mm}
	\subfigure[Attenuation] {\includegraphics[width=.47\linewidth]{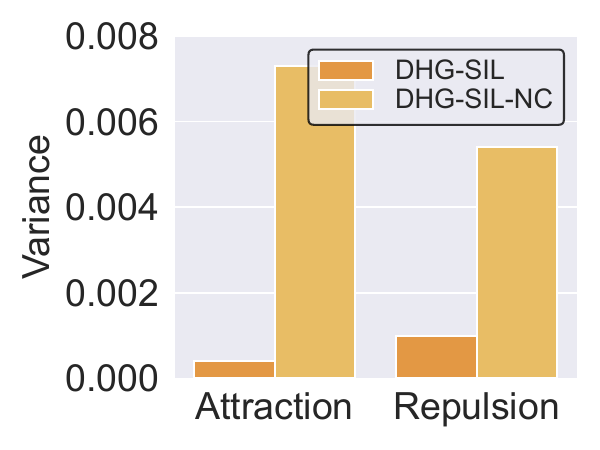}}
	\caption{Effect on contrastive learning strategies for attraction and repulsion interpretability.}
	\label{fig:contrastive}
\end{figure}

\subsection{Effect on Contrastive Learning Strategies}

Here, we evaluate the effect of our proposed contrastive optimization by comparing DHG-SIL with a variant of DHG-SIL named \textbf{DHG-SIL-NC}, which is without contrastive losses in optimization. We calculate the variance of each city across $24$ time steps and calculate the average of all cities in the YRD dataset to describe the interpretability brought by the contrastive loss. Because for each city, the time-varying values of attraction and repulsion across months should not usually drastically fluctuate. Figure~\ref{fig:contrastive} demonstrates that our contrastive loss significantly stabilizes~(i.e., decreases the variance) the optimized values of interpretable variables, i.e., improves the robustness of interpretability of DHG-SIL.

\begin{figure}[t]
    \centering
    \subfigure[Attraction - Inflow] {\includegraphics[width=.47\linewidth]{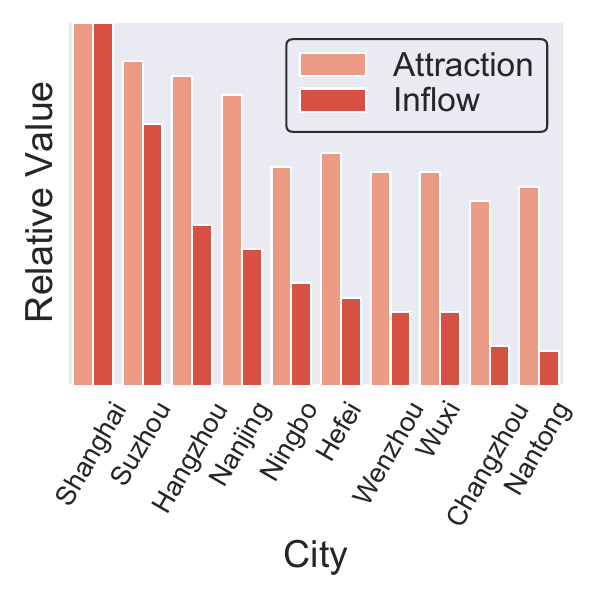}\label{fig:attract_inflow}}
    \hspace{3mm}
    \subfigure[Repulsion - Outflow]{\includegraphics[width=.47\linewidth]{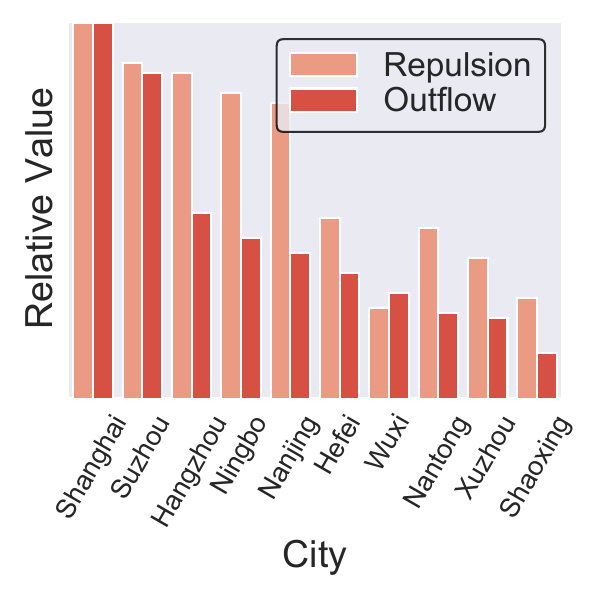}\label{fig:repulse_outflow}}
    \caption{The attraction and repulsion of top $10$ cities with the highest inflow and outflow.}
    \label{fig:variable_analysis}
\end{figure}

\subsection{Interpretable Variables Analysis}

Our interpretive learning framework introduces four variables to capture urban labor migration dynamics: attraction intensity ($\alpha_\delta$), attraction attenuation ($\sigma_\delta$), repulsion intensity ($\alpha_\mu$), and repulsion attenuation ($\sigma_\mu$). The incorporation of spatial context transcends traditional methods that only focus on migration inflow and outflow.
In assessing these variables' geographic relevance, we filter the top ten cities by attraction/repulsion intensity within the YRD dataset, correlating these with migration statistics as depicted in Figure~\ref{fig:variable_analysis}. Figure~\ref{fig:attract_inflow} shows a pronounced correlation between attraction intensity and inflow, albeit with anomalies such as Ningbo and Changzhou; notably, $78.50\%$ of migrants to Ningbo originate from within the city, surpassing the self-inflow rates of Nanjing ($57.81\%$) and Hefei ($65.23\%$). Conversely, Figure~\ref{fig:repulse_outflow} indicates a similar trend for repulsion intensity and outflow, with Wuxi as an outlier where merely $4.44\%$ and $2.72\%$ of migrants move to Shanghai and Nanjing, respectively, compared to Nantong's $6.52\%$ and $3.16\%$ despite its lower overall outflow. These examples and statistics substantiate that our framework enriches labor migration analysis by integrating geographic information, offering a more sophisticated portrayal of urban migration phenomena.

\begin{figure}[t]
	\centering
	\subfigure[YRD] {\includegraphics[width=.32\linewidth]{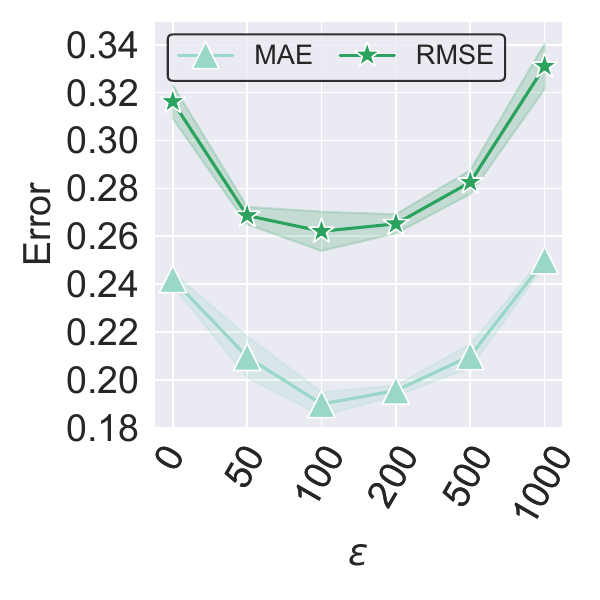}}
	\subfigure[PRD] {\includegraphics[width=.32\linewidth]{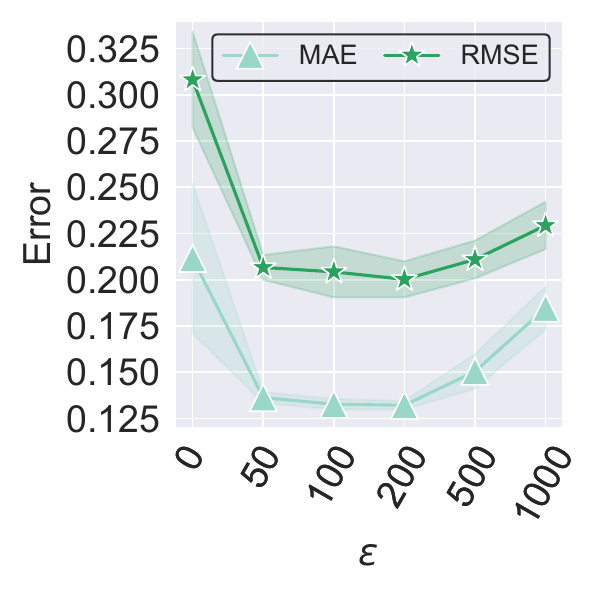}}
	\subfigure[BTH] {\includegraphics[width=.32\linewidth]{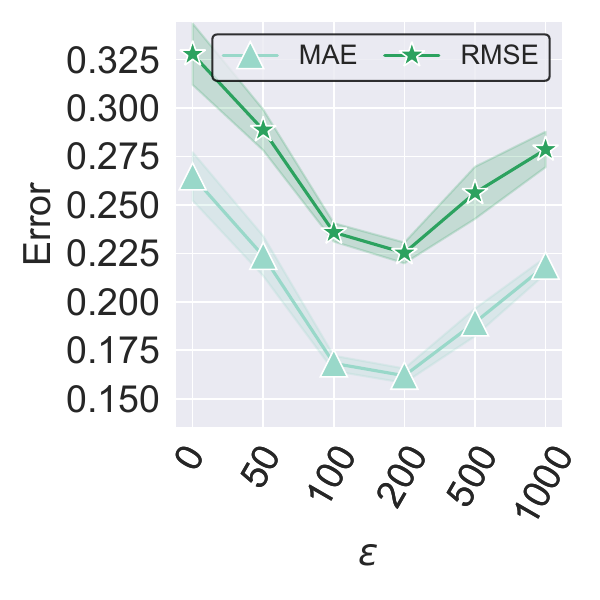}}
	\caption{Effect on different geography-homophilic threshold for different urban agglomerations.}
	\label{fig:max_distance}
\end{figure}

\subsection{Effect on Geography-Homophilic Distance Threshold}

We study the distance thresholds to enhance prediction accuracy and provide insights into urban agglomeration in Figure~\ref{fig:max_distance}. We vary $\epsilon$ from $0$km to $1000$km and observe that DHG-SIL underperforms when $\epsilon$ is set too low or too high, leading to disjointed or oversmooth message passing in GNN, respectively.
Moreover, urban agglomerations have distinct geographic homophily. For example, cities in YRD are more diverse among its large area, leading to a lower best $\epsilon$. While PRD has only $9$ close cities that are almost geography-homophilic, i.e., no negative performance influence with a high $\epsilon$.

\begin{figure}[t]
	\centering
	\subfigure[Attraction distribution] {\includegraphics[width=.47\linewidth]{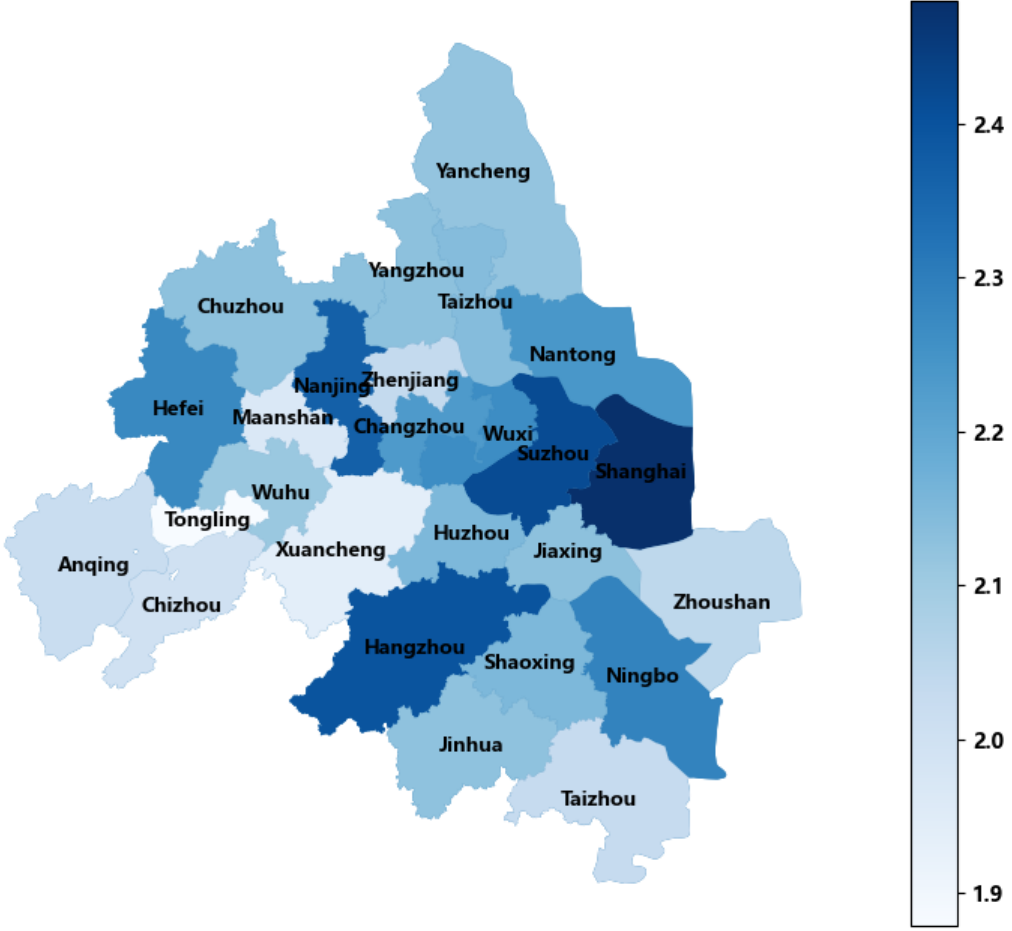}}
	\hspace{3mm}
	\subfigure[Repulsion distribution] {\includegraphics[width=.47\linewidth]{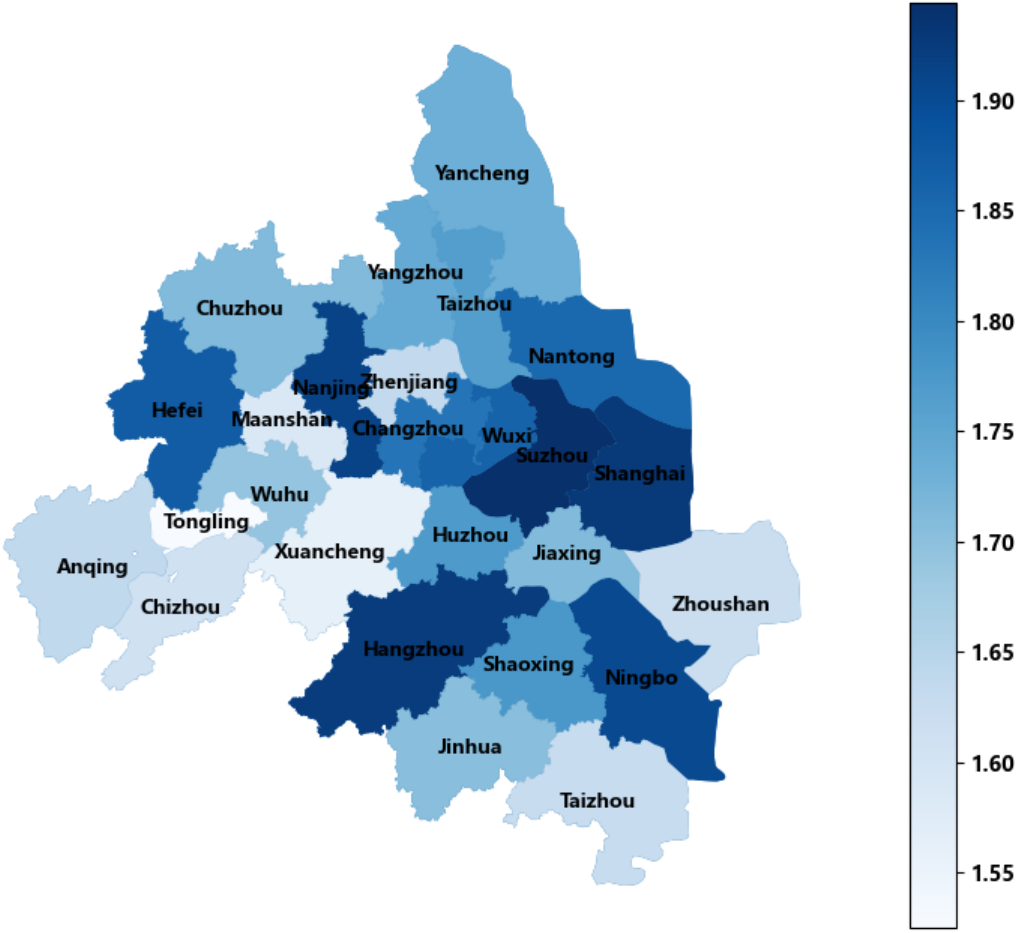}}
	\caption{The attraction and repulsion visualization of Yangtze River Delta Urban Agglomerations.}
	\label{fig:at_tr_dist}
\end{figure}

\subsection{Regional Distribution Visualization}

This study takes YRD as an example to examine the interpretable city characteristics including attraction and repulsion intensity by creating geographical visualization in Figure~\ref{fig:at_tr_dist}. The eastern part is found to be the most attractive area for laborers, while the northern and western parts attract fewer laborers. Comparing attraction and repulsion, we observe that the repulsion of less attractive areas is more intensive than their attraction because laborers here have more intention toward more developed cities than local occupation.
In conclusion, beyond precise prediction of future migration trends, we can analyze them from a spatial perspective for knowledge discovery based on our optimized variables.

\subsection{System Deployment}

\begin{figure}[t]
    \centering
    \includegraphics[width=\linewidth]{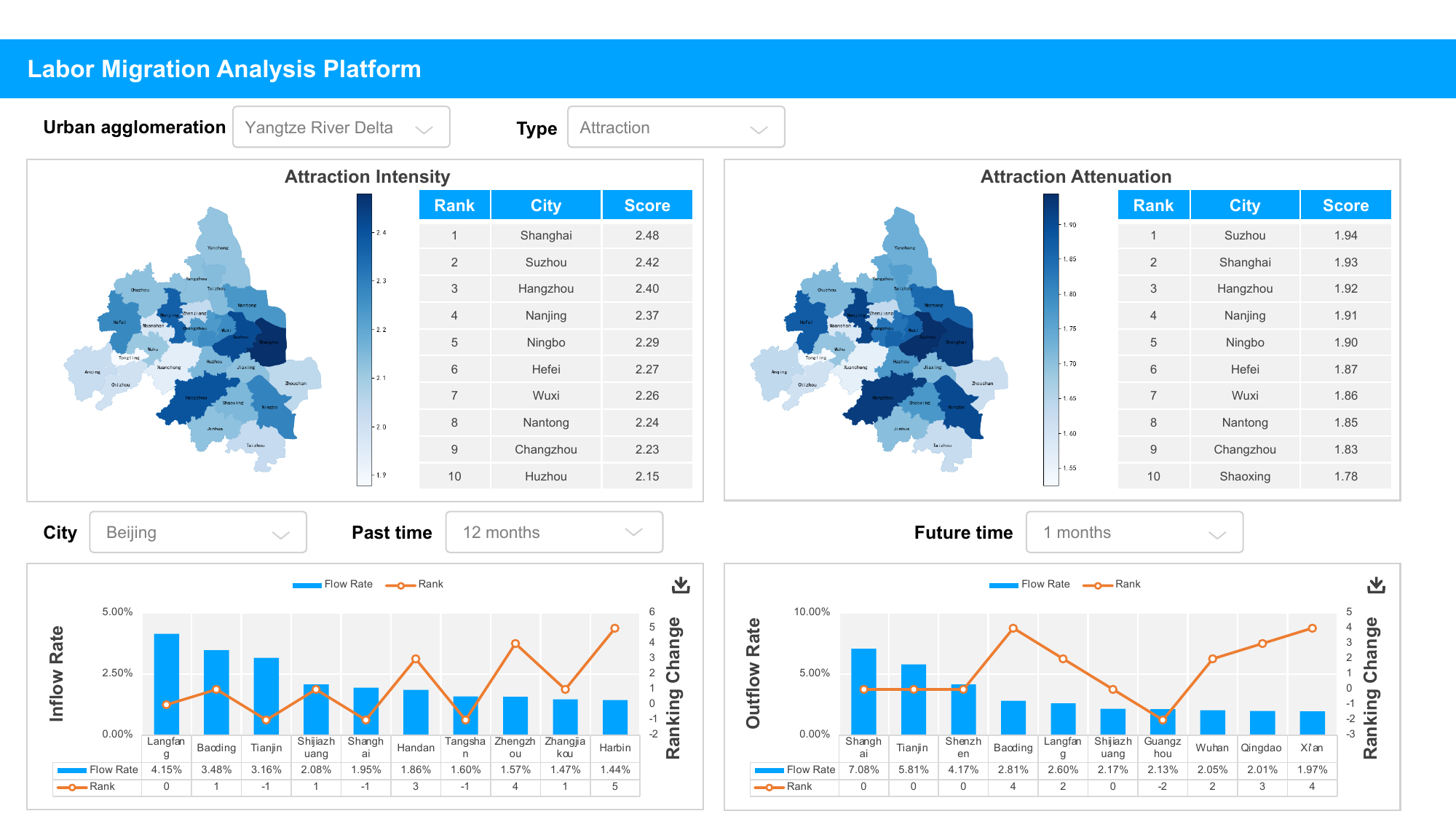}
    \caption{Deployed system.}
    \label{fig:deployment}
\end{figure}

DHG-SIL has been integrated as a core component of our partner's intelligent human resource system, providing an in-depth analysis of city labor migration. The system design in Figure~\ref{fig:deployment} displays four variables of urban agglomeration, offering insight into regional attraction and repulsion statistics. It also shows future inflow and outflow trends of selected cities. Additionally, the system supports reports on urban talent attraction in China, utilizing large-scale job query data in a popular search engine. This report has gained significant attention in Chinese society since its publication.

\section{Related Works}

\textbf{Labor Migration Modeling.}
Labor migration studies have spanned over several decades with an initial emphasis on the economic impact~\cite{stark1985new}. Early investigations utilized mathematical models to characterize migration dynamics and network analysis to elucidate migration traits~\cite{zhao2003role}. These foundational sociological inquiries inform contemporary approaches, including granular modeling and urban feature mapping from geographical and mobility standpoints.
Contemporary research typically employs empirical mobility data~\cite{rinken2022leveraging}. Case studies, such as household surveys in western Mexico, have elucidated perceptions of migratory opportunities~\cite{cornelius2019labor}. Yet, data accessibility challenges due to privacy concerns and technical constraints have spurred simulated migration studies using surrogate datasets~\cite{tjaden2021measuring,gao2020does}.
Web search analytics, particularly from Bing.com, have recently been harnessed to quantify migration trends~\cite{lin2019forecasting,chancellor2018measuring}. Furthermore, European labor migration research has leveraged data from LinkedIn, Eurostat, and job market websites to validate traditional statistical hypotheses~\cite{mamertino2016online,mamertino2019migration}.
Different from previous studies, this study pioneers the use of deep learning with interpretable domain variables, utilizing extensive job search data to uncover the underlying mechanisms of labor migration behavior.

\noindent
\textbf{GNN for Graph Homophiliy.}
Homophily is a foundational assumption in GNN design, yet real-world networks often exhibit heterophily, necessitating tailored GNN frameworks for practical graph learning tasks~\cite{zheng2022graph}.
High-order neighbors mixing is a popular research direction for heterophilic graphs. MixHop~\cite{abu2019mixhop} and H2GCN~\cite{zhu2020beyond} are representative methods that aggregate two-hop neighborhoods in message passing.
Moreover, UGCN~\cite{jin2021universal} further restricts the two-hop neighbor set with a strong connection.
Instead of extending the subgraph from the ego node within $k$-hop, some works try to find potential neighbors with metrics or mechanisms, such as the 2D Euclidean geometry location~\cite{pei2020geom}, attention~\cite{liu2021non,yang2022graph} and homophily degree matrix~\cite{wang2022powerful}.
Conventional approaches for preserving graph homophily typically focus on a single dimension. In our study, we improve existing GNN by concurrently addressing two distinct aspects of homophily within a feature set to offer a fresh perspective of learning on heterophilic graphs.

\section{Conclusion}

In this paper, we investigated labor migration modeling based on the large-scale job query data collected from one of the largest web search engines. As the first deep learning and big data-driven attempt on this topic, our work proposed a Discrepant Homophily co-preserved Graph enhanced Sequential Interpretable Learning~(DHG-SIL) framework to forecast the cross-city time-varying labor migration trend. Specifically, we first devised the Discrepant Homophily co-preserved Graph Convolutional Network~(DH-GCN) to incorporate the city feature with joint consideration of preserving geography- and mobility- homophily, which are naturally discrepant, through operating a novel graph convolution kernel on a constructed geography graph. Then four interpretable variables describing the city labor migration properties were co-optimized with city representations for absorbing knowledge from overall prediction optimization. 
Afterward, we leveraged a temporal model to encode the labor migration series for building sequential dependencies. We also designed contrastive losses to enhance interpretable variables by mining hidden supervision information. Extensive experiments on quantitative and qualitative studies concurrently showed the advancement of DHG-SIL, which also provided knowledge for various downstream tasks, such as governmental decisions on labor policies and the proper selection of industries.
Our proposed DHG-SIL has deployed an intelligent human resource system of a cooperative partner, and the system supports the China city talent attraction reports in the past years.



\bibliographystyle{ACM-Reference-Format}
\bibliography{reference}


\end{document}